\documentclass[twoside,11pt]{article}

\usepackage{jmlr2e}

\usepackage[utf8]{inputenc}
\usepackage[T1]{fontenc}
\usepackage{amsmath,amsfonts}
\usepackage{xcolor}
\usepackage{booktabs}
\usepackage{listings}
\usepackage{subcaption}
\usepackage{multirow}
\usepackage{array}
\usepackage{float}
\usepackage{enumitem}
\usepackage[section]{placeins}   
\usepackage{url}
\hypersetup{hidelinks}

\graphicspath{{images/}}

\newcommand{\mosaic}{\textsc{Mosaic}}
\newcommand{\eg}{\emph{e.g.}}

\newcommand{\authormail}[1]{\mdseries\footnotesize\textcolor{gray}{\textsc{#1}}}

\newcommand{\yes}{\textcolor{green!60!black}{$\checkmark$}}
\newcommand{\no}{\textcolor{red}{$\times$}}
\newcommand{\partialsup}{\textcolor{blue}{$\circ$}}

\usepackage{lastpage}
\jmlrheading{26}{2026}{1-\pageref{LastPage}}{[DATE SUBMITTED]}{[DATE PUBLISHED]}{[PAPER ID]}{Abdulhamid M. Mousa, Ming Liu, Rakhmonberdi Khajiev, Jalaledin M. Azzabi, Abdulkarim M. Mousa, Peng Yong, Yunusa Haruna, and Jinhui Pang}

\ShortHeadings{MOSAIC: A Universal Interface for Cross-Paradigm Agent Mixing}{Mousa, Yong, and Liu}

\firstpageno{1}

\begin{document}

\title{MOSAIC: A Universal Agent-Level Interface for Cross-Paradigm Agent Mixing and Human-AI Collaboration}

\author{%
  \begin{tabular*}{\textwidth}{@{}l@{\extracolsep{\fill}}r@{}}
  Abdulhamid M.~Mousa$^{1}$\thanks{Abdulhamid M. Mousa, Abdulkarim M. Mousa, and Peng Yong are the main developers. They contributed equally to this paper.}
      & \authormail{mousa.abdulhamid@bit.edu.cn}         \\
  Ming Liu$^{1}$
      & \authormail{bit411liu@bit.edu.cn}                \\
  Rakhmonberdi Khajiev$^{1}$
      & \authormail{khajiev.rakhmonberdi@bit.edu.cn}     \\
  Jalaledin M.~Azzabi$^{1}$
      & \authormail{jalaledin.azzabi@bit.edu.cn}         \\
  Abdulkarim M.~Mousa$^{3}$\footnotemark[1]
      & \authormail{abdulkarim.mousa@outlook.com}        \\
  Peng Yong$^{1}$\footnotemark[1]
      & \authormail{m18801320372@163.com}                \\
  Yunusa Haruna$^{2}$
      & \authormail{yunusa2k2@buaa.edu.cn}               \\
  Jinhui Pang$^{4}$\thanks{Corresponding author.}
      & \authormail{pangjinhui@bit.edu.cn}               \\
  \end{tabular*}\\[8pt]
  {\footnotesize\mdseries\itshape\textcolor{gray}%
    {$^{1}$School of Optics and Photonics, Beijing Institute of Technology, Beijing 100081, China}}\\
  {\footnotesize\mdseries\itshape\textcolor{gray}%
    {$^{2}$School of Automation Science and Electrical Engineering, Beihang University, Beijing 100191, China}}\\
  {\footnotesize\mdseries\itshape\textcolor{gray}%
    {$^{3}$Mathematics Department, Faculty of Science, Ain Shams University, Cairo, Egypt}}\\
  {\footnotesize\mdseries\itshape\textcolor{gray}%
    {$^{4}$School of Computer Science and Technology, Beijing Institute of Technology, Beijing 100081, China}}\\[4pt]
}

\editor{[ACTION EDITOR NAME]}

\maketitle

\begin{abstract}%
Existing infrastructure cannot deploy agents from different decision-making paradigms within the same environment, making fair cross-paradigm comparison under identical conditions impossible. We present \mosaic{}, an open-source platform that enables heterogeneous agents (RL policies, LLMs, VLMs, and human operators) to act within shared reinforcement learning environments in ad-hoc team settings with reproducible results. \mosaic{} introduces three contributions. (i)~An \emph{IPC-based worker protocol} that wraps native and third-party frameworks as isolated subprocess workers, each executing its own training and inference logic unmodified and communicating through a versioned inter-process protocol. (ii)~An \emph{operator abstraction} that forms an agent-level interface by mapping workers to agent slots: each operator, regardless of whether it is backed by an RL policy, an LLM, or a human, conforms to a minimal universal interface. (iii)~A \emph{deterministic cross-paradigm evaluation} framework with two complementary modes: a \emph{manual mode} that advances up to $N$ operators in lock-step under shared seeds for fine-grained visual inspection of behavioural differences; and a \emph{script mode} that drives automated, long-running evaluation via declarative Python scripts for reproducible experiments. Our documentation is released at: \textbf{https://mosaic-platform.readthedocs.io}.
\end{abstract}

\begin{keywords}
Reinforcement Learning, Large Language Models, Multi-Agent Systems,
Agent-Level Interface, Cross-Paradigm Evaluation
\end{keywords}

\section{Introduction}
\label{sec:intro}

Reinforcement learning (RL) frameworks
(RLlib~\cite{pmlr-v80-liang18b}; CleanRL~\cite{DBLP:journals/jmlr/HuangDYBCMA22};
 Stable-Baselines3~\cite{10.5555/3546258.3546526}; Tianshou~\cite{10.5555/3586589.3586856};
 XuanCe~\cite{liu2023xuancecomprehensiveunifieddeep}; OpenRL~\cite{huang2023openrlunifiedreinforcementlearning};
 Acme~\cite{hoffman2022acmeresearchframeworkdistributed})
and LLM/VLM benchmarks
(BALROG~\cite{paglieri2025balrogbenchmarkingagenticllm}; AgentBench~\cite{liu2025agentbenchevaluatingllmsagents};
 TextArena~\cite{guertler2025textarena}; GameBench~\cite{costarelli2024gamebenchevaluatingstrategicreasoning};
 AgentGym~\cite{xi2024agentgymevolvinglargelanguage})
have matured independently.
Gymnasium~\cite{towers2025gymnasiumstandardinterfacereinforcement} and
PettingZoo~\cite{terry2021pettingzoogymmultiagentreinforcement} standardised the \emph{environment} side of
the agent/environment loop, enabling any algorithm to connect to any compatible
simulator.
However, the \emph{agent} side remains fragmented: RL trainers expect tensor
observations and produce integer actions, LLM agents expect text prompts and produce
text responses, and human operators need interactive interfaces.
\mosaic{} addresses this fragmentation through a universal agent-level interface that
decouples an environment's agent slots from the computational paradigm of the
decision-makers controlling them.
Table~\ref{tab:comparison} (Appendix~\ref{app:comparison}) surveys 23 existing
frameworks across these dimensions.

\section{Related Work}
\label{sec:related}

Existing reinforcement-learning libraries such as
RLlib~\cite{pmlr-v80-liang18b},
CleanRL~\cite{DBLP:journals/jmlr/HuangDYBCMA22},
Stable-Baselines3~\cite{10.5555/3546258.3546526},
Tianshou~\cite{10.5555/3586589.3586856},
XuanCe~\cite{liu2023xuancecomprehensiveunifieddeep},
OpenRL~\cite{huang2023openrlunifiedreinforcementlearning},
Acme~\cite{hoffman2022acmeresearchframeworkdistributed},
BenchMARL~\cite{bettini2024benchmarl}, and
JaxMARL~\cite{DBLP:conf/nips/RutherfordEG0LI24}
provide strong support for training, benchmarking, and reproducible evaluation
within the RL paradigm.
In parallel, BALROG~\cite{paglieri2025balrogbenchmarkingagenticllm},
AgentBench~\cite{liu2025agentbenchevaluatingllmsagents},
TextArena~\cite{guertler2025textarena},
GameBench~\cite{costarelli2024gamebenchevaluatingstrategicreasoning},
AgentGym~\cite{xi2024agentgymevolvinglargelanguage}, and related benchmarks
evaluate LLM/VLM agents in reinforcement learning environments, while
Overcooked-AI~\cite{carroll2020utilitylearninghumanshumanai} and
CREW~\cite{zhang2025crewfacilitatinghumanaiteaming} study human--AI collaboration.
These systems largely remain specialised by decision-making paradigm or application
domain.
\mosaic{} instead targets the agent-interface boundary: RL policies, LLMs, VLMs,
humans, and scripted agents are exposed through an operator abstraction and can be
deployed in the same environment under a shared evaluation protocol.
This also differs from ad hoc teamwork~\cite{stone2010adhoc,mirsky2022survey} and
zero-shot coordination~\cite{wang2024naht,wang2024zsceval}, which study cooperation
with unfamiliar teammates but generally assume compatible observation and action
interfaces.
\mosaic{} focuses on cross-paradigm heterogeneity, where teammates may use different
observation modalities, action representations, and decision processes.

\section{Software Design}
\label{sec:design}

\begin{figure}[ht]
    \centering
    \includegraphics[width=\textwidth]{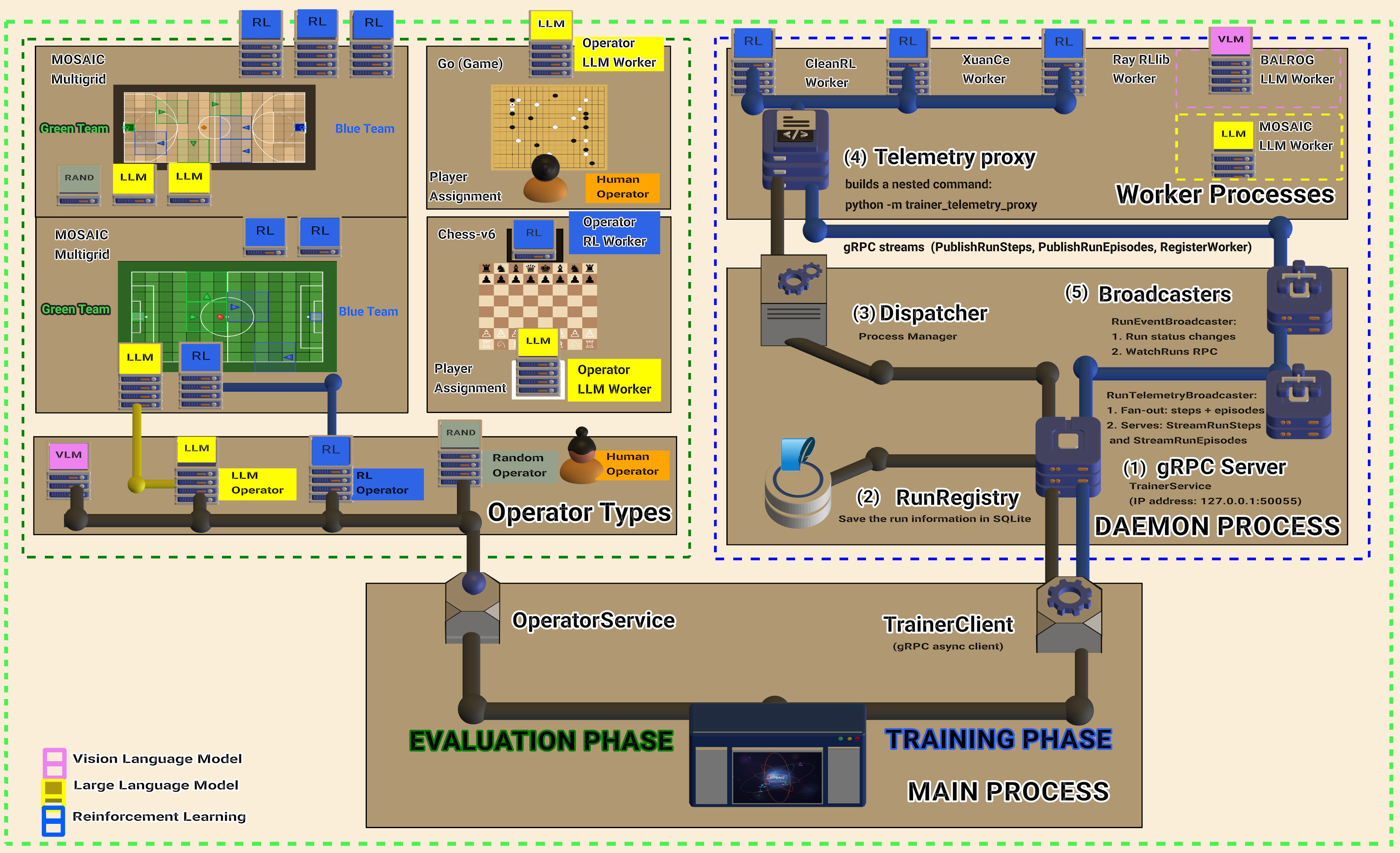}
    \caption{\mosaic{} architecture.
    \textbf{Left:} Operator types (RL, LLM, VLM, Human, Random) deployed across
    environments.
    \textbf{Right:} Internal process structure: Daemon (gRPC, RunRegistry, Dispatcher),
    Worker Processes (CleanRL, XuanCe, RLlib, BALROG, MOSAIC~LLM),
    Telemetry Proxy, and Qt6 Main Process.}
    \label{fig:architecture}
\end{figure}

\mosaic{} follows a three-tier architecture separating
\emph{orchestration} (Qt6 GUI),
\emph{communication} (IPC protocol), and
\emph{execution} (worker subprocesses), as shown in Figure~\ref{fig:architecture}.

\paragraph{Orchestration layer.}
The Trainer Daemon acts as the authoritative control plane: it spawns and supervises
workers as isolated subprocesses via \texttt{os.setsid()} for process-group isolation,
establishes bidirectional IPC with a structured capability handshake, routes commands
(\texttt{reset}, \texttt{step}, \texttt{train}), aggregates live telemetry into
SQLite-backed models, and exposes pause/resume controls.
The Qt6 main process is the user-facing interface that issues operator commands to the
Daemon via gRPC; it never embeds algorithmic logic.

\paragraph{Worker protocol.}
Each worker subprocess communicates via a lightweight JSON protocol over stdin/stdout.
Commands flow from the GUI:
\texttt{\{"cmd":"reset","seed":42\}},
\texttt{\{"cmd":"step"\}},
\texttt{\{"cmd":"stop"\}}.
Typed responses return:
\texttt{ready} (environment metadata, seed, observation shape after reset),
\texttt{step} (action, reward, terminated, render payload),
\texttt{episode\_done} (total reward, episode length),
and \texttt{error} (failure message).
In batch mode, workers accept CLI arguments and emit JSONL to stdout; a telemetry proxy
(sidecar process) parses JSON lines, validates against versioned schemas, converts them
to Protocol Buffer messages, and forwards them to the daemon via gRPC streams
(\texttt{PublishRunSteps}, \texttt{PublishRunEpisodes}).

\paragraph{Operator abstraction.}
An \emph{operator} maps one or more workers to agent slots in an environment.
The \texttt{OperatorLauncher} selects the appropriate worker based on operator type:
RL operators invoke framework workers (CleanRL, XuanCe, Ray RLlib) with
\texttt{{-}{-}interactive} flags;
LLM operators route to environment-specific workers: BALROG for single-agent MiniGrid/BabyAI,
or the native MOSAIC LLM Worker for multi-agent coordination and adversarial setups;
human operators connect keyboard input via the Human Worker for human-in-the-loop evaluation;
and baseline operators invoke the Random Worker with random, noop, or cycling behaviors.
For multi-agent environments, \texttt{MultiAgentOperatorHandle} manages one worker process
per agent, routing per-agent \texttt{select\_action} commands and aggregating responses.

\begin{figure}[h]
\centering
\includegraphics[width=\textwidth]{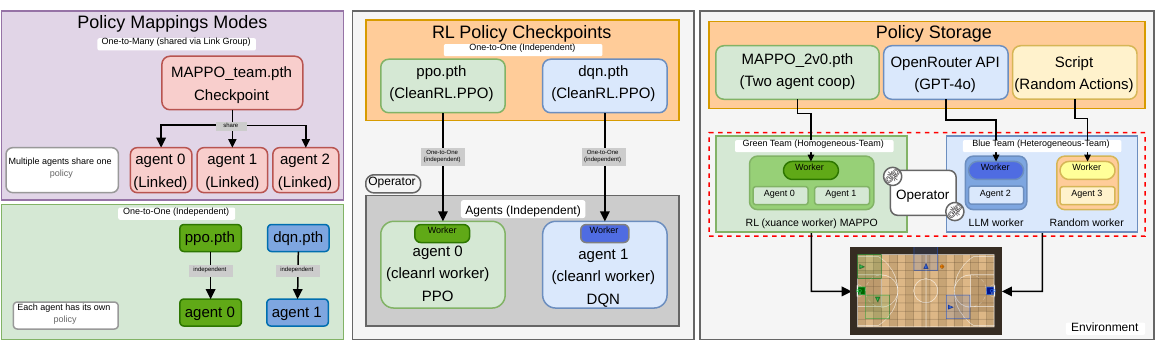}
\caption{Operator abstraction: how different agent types (RL, LLM, human, and baseline)
are mapped to environment agent slots through the \texttt{OperatorLauncher} and
\texttt{MultiAgentOperatorHandle}.}
\label{fig:operator-abstraction}
\end{figure}

\paragraph{Policy mappings.}
\mosaic{} offers two assignment modes. \emph{One-to-one} gives each slot its own
worker and policy, so RL, LLM, Human, and Random agents can coexist freely within
the same environment. \emph{One-to-many} binds agents via a \texttt{LinkGroup}:
one primary agent holds the checkpoint path and all linked agents inherit it
automatically. This is the standard setup for MAPPO and IPPO, where all slots share
a single policy file; updating the primary path propagates instantly to every linked
agent. The two modes combine freely to form arbitrary heterogeneous teams.

\section{Software Quality and Availability}
\label{sec:quality}

\paragraph{Documentation.}
The documentation site (\url{https://mosaic-platform.readthedocs.io}) comprises
155+ pages covering: installation guides for Ubuntu, Windows, and WSL with common-error
troubleshooting; quickstart tutorials; per-environment guides for all 33 families;
full architecture documentation; API reference (Operator, Worker, Environment);
a contributing guide; and a changelog.

\paragraph{License and availability.}
Source code: \url{https://github.com/Abdulhamid97Mousa/MOSAIC}.
Documentation: \url{https://mosaic-platform.readthedocs.io}.

\section{Conclusion}

We presented \mosaic{}, an open-source platform that standardizes the agent side of the
agent/environment interface, complementing Gymnasium's environment standardisation.
Through the Operator Protocol, IPC-based worker isolation, and deterministic
cross-paradigm evaluation, \mosaic{} enables the first infrastructure for fair,
reproducible comparison between RL, LLM/VLM, and human decision-makers in shared
multi-agent environments.

\newpage

\acks{The authors thank the Beijing Institute of Technology for computing resources.
We acknowledge the developers of Gymnasium, PettingZoo, CleanRL, XuanCe, Ray~RLlib,
and BALROG, whose open-source contributions made this work possible.}

\appendix

\section{Comparison with Existing Frameworks}
\label{app:comparison}

\begingroup
\setlength{\tabcolsep}{3pt}
\footnotesize
\captionsetup{hypcap=false}
\begin{table}[htbp]
\centering
\caption{Comparison with 23 existing frameworks.
\textit{Agent Paradigms}: supported decision-maker types.
\textit{Framework}: third-party algorithms integrate without source-code modifications.
\textit{GUI}: real-time visualisation during execution.
\textit{Cross-Paradigm}: infrastructure for comparing different agent types
(\eg, RL, LLM, VLM, Human) on identical or different environments with shared seeds.
\yes~Supported, \no~Not supported, \partialsup~Partial.}
\label{tab:comparison}
\resizebox{\textwidth}{!}{%
\begin{tabular}{@{}l cccc c cc@{}}
\toprule
& \multicolumn{4}{c}{{\footnotesize\textbf{Agent Paradigms}}}
& \multicolumn{2}{c}{{\footnotesize\textbf{Infrastructure}}}
& {\footnotesize\textbf{Evaluation}} \\
\cmidrule(lr){2-5}\cmidrule(lr){6-7}\cmidrule(lr){8-8}
{\footnotesize\textbf{System}}
  & {\footnotesize\textbf{RL}} & {\footnotesize\textbf{LLM}} & {\footnotesize\textbf{VLM}} & {\footnotesize\textbf{H}}
  & {\footnotesize\textbf{Framework}} & {\footnotesize\textbf{GUI}}
  & {\footnotesize\textbf{Cross-Paradigm}} \\
\midrule
\multicolumn{8}{@{}l}{\textit{RL Frameworks}} \\[1pt]
RLlib \cite{pmlr-v80-liang18b}
  & \yes & \no & \no & \no & \yes & \no & \no \\
CleanRL \cite{DBLP:journals/jmlr/HuangDYBCMA22}
  & \yes & \no & \no & \no & \yes & \no & \no \\
Tianshou \cite{10.5555/3586589.3586856}
  & \yes & \no & \no & \no & \yes & \no & \no \\
Acme \cite{hoffman2022acmeresearchframeworkdistributed}
  & \yes & \no & \no & \no & \yes & \no & \no \\
XuanCe \cite{liu2023xuancecomprehensiveunifieddeep}
  & \yes & \no & \no & \no & \yes & \no & \no \\
OpenRL \cite{huang2023openrlunifiedreinforcementlearning}
  & \yes & \no & \no & \no & \yes & \no & \no \\
Stable-Baselines3 \cite{10.5555/3546258.3546526}
  & \yes & \no & \no & \no & \yes & \no & \no \\
Coach \cite{caspi_itai_2017_1134899}
  & \yes & \no & \no & \no & \yes & \yes & \no \\
BenchMARL \cite{bettini2024benchmarl}
  & \yes & \no & \no & \no & \yes & \no & \no \\
JaxMARL \cite{DBLP:conf/nips/RutherfordEG0LI24}
  & \yes & \no & \no & \no & \yes & \no & \no \\
Overcooked-AI \cite{carroll2020utilitylearninghumanshumanai}
  & \yes & \no & \no & \yes & \no & \yes & \no \\
\midrule
\multicolumn{8}{@{}l}{\textit{LLM/VLM Benchmarks}} \\[1pt]
BALROG \cite{paglieri2025balrogbenchmarkingagenticllm}
  & \no & \yes & \yes & \no & \yes & \no & \no \\
TextArena \cite{guertler2025textarena}
  & \no & \yes & \no & \yes & \yes & \no & \no \\
GameBench \cite{costarelli2024gamebenchevaluatingstrategicreasoning}
  & \no & \yes & \no & \no & \yes & \no & \no \\
lmgame-Bench \cite{hu2025lmgamebenchgoodllmsplaying}
  & \no & \yes & \no & \no & \yes & \no & \no \\
LLM Chess \cite{kolasani2025llmchessbenchmarkingreasoning}
  & \yes & \yes & \no & \no & \yes & \no & \no \\
LLM-Game-Bench \cite{topsakal2024evaluatinglargelanguagemodels}
  & \no & \yes & \no & \no & \yes & \partialsup & \no \\
AgentBench \cite{liu2025agentbenchevaluatingllmsagents}
  & \no & \yes & \no & \no & \yes & \no & \no \\
MultiAgentBench \cite{zhu2025multiagentbenchevaluatingcollaborationcompetition}
  & \no & \yes & \no & \no & \yes & \no & \no \\
GAMEBoT \cite{lin-etal-2025-gamebot}
  & \no & \yes & \no & \no & \yes & \no & \no \\
Collab-Overcooked \cite{sun-etal-2025-collab}
  & \partialsup & \yes & \no & \no & \yes & \no & \no \\
BotzoneBench \cite{li2026botzonebenchscalablellmevaluation}
  & \no & \yes & \no & \no & \yes & \no & \no \\
AgentGym \cite{xi2024agentgymevolvinglargelanguage}
  & \no & \yes & \no & \no & \yes & \no & \no \\
\midrule
\textbf{MOSAIC (Ours)}
  & \yes & \yes & \yes & \yes & \yes & \yes & \yes \\
\bottomrule
\end{tabular}}
\end{table}
\endgroup

Table~\ref{tab:comparison} surveys 23 systems across seven properties. The most
immediate observation is the paradigm split: all eleven RL frameworks (RLlib through
Overcooked-AI) restrict their agent-paradigm support to RL and, in one case, Human,
while the thirteen LLM/VLM benchmarks (BALROG through AgentGym) cover language
agents exclusively, with no RL training or human-in-the-loop infrastructure. GUI
support is rare even within each paradigm, appearing only in Coach and Overcooked-AI
among RL frameworks and partially in LLM-Game-Bench among LLM benchmarks. Most
importantly, the \emph{Cross-Paradigm} column is unchecked for every prior system:
none provides the infrastructure needed to run RL, LLM, VLM, and human decision-makers
side-by-side on the same environment under shared random seeds. \mosaic{} is the only
system to check all seven columns simultaneously.

\section{Usage Examples}
\label{app:usage}

\paragraph{Installation.}
\mosaic{} uses a modular extras system so that users install only the workers and
environments they need:

\begin{lstlisting}[language=bash,numbers=none]
pip install -e ".[cleanrl,minigrid]"        # RL + grid worlds
pip install -e ".[xuance,mosaic_multigrid]"  # MARL + soccer
pip install -e ".[full]"                     # everything
\end{lstlisting}

\paragraph{Configuring heterogeneous agents.}
A single configuration assigns different decision-makers to each agent slot via
\texttt{WorkerAssignment}, while \texttt{LinkGroup} enables agents to share a
single policy checkpoint (essential for MAPPO/IPPO where all agents' parameters
reside in one file).
The following excerpt deploys a trained MAPPO policy ($\bar{\pi}^{RL}$) and a
GPT-4o agent ($\lambda^{LLM}$) as teammates against a second RL agent and a
random baseline ($\rho$) in 2v2 soccer:

\begin{lstlisting}[language=python,numbers=none]
config = OperatorConfig.multi_agent(
    operator_id="heterogeneous_team",
    env_name="multigrid",
    task="MosaicMultiGrid-Soccer-2vs2-IndAgObs-v0",
    player_workers={
        "green_0": WorkerAssignment(
            worker_id="xuance_worker",
            worker_type="rl",
            settings={"algorithm": "mappo"}),
        "green_1": WorkerAssignment(
            worker_id="mosaic_llm_worker",
            worker_type="llm",
            settings={"model_id": "gpt-4o",
                       "temperature": 0}),
        "blue_0": WorkerAssignment(
            worker_id="xuance_worker",
            worker_type="rl",
            settings={"algorithm": "mappo"}),
        "blue_1": WorkerAssignment(
            worker_id="random_worker",
            worker_type="random", settings={}),
    },
    link_groups={
        "link_0": LinkGroup(
            primary_agent="green_0",
            linked_agents=["blue_0"],
            policy_path="mappo_2v2.pth",
            algorithm="mappo"),
    },
)
\end{lstlisting}

The \texttt{LinkGroup} binds \texttt{green\_0} and \texttt{blue\_0} to the same
MAPPO checkpoint; updating the primary agent's path propagates automatically.
This configuration demonstrates heterogeneous ad-hoc teamwork: a frozen RL
policy is paired with an LLM teammate in 2v2, isolating the cross-paradigm
variable.

\section{FastLane: Zero-Serialisation Live Visualisation}
\label{app:fastlane}

FastLane is \mosaic{}'s \textbf{zero-serialisation live visualisation path}. It streams
live RGB frames from a training worker subprocess directly into POSIX shared memory,
bypassing the slow-lane gRPC/SQLite pipeline entirely. The GUI-side
\texttt{FastLaneConsumer} polls the buffer every 16 ms ($\approx$60 Hz) and renders
the latest frame via a Qt Quick \texttt{FastLaneTab}. Unlike existing approaches that either render in-process (blocking training) or stream
via network sockets (NVIDIA sim-web-visualizer), FastLane fully decouples
visualisation from the training loop: the worker writes frames without waiting for the
GUI, and the GUI reads without blocking the worker. Prior shared-memory systems in RL
(OpenAI Baselines \texttt{ShmemVecEnv}~\cite{dhariwal2017openai},
Sample Factory~\cite{petrenko2020sample}, EnvPool~\cite{weng2022envpool},
TorchRL~\cite{bou2023torchrl}) all transfer training data between workers. FastLane is
the first to apply shared memory to \textbf{visualisation output}. The core is a Single-Producer Single-Consumer (SPSC) ring buffer in POSIX shared
memory, inspired by the LMAX Disruptor~\cite{thompson2011disruptor}. Writer and reader
coordinate via an odd/even sequence-number protocol (seqlock pattern): the writer marks
a slot odd (write in progress), copies the RGB payload, then marks it even (committed);
the reader skips odd slots and retries on seq-mismatch (torn read). Correctness on
x86 relies on Total Store Order plus the seq-mismatch retry, not the GIL. The
benchmarks below were run on Ubuntu 22.04 (x86-64, ERYING Polestar Z790, CPython 3.11)
with all tests passing and zero torn reads across 155,000 frames.

\subsection{Empirical Validation}

The following benchmarks were run on Ubuntu 22.04 (x86-64, ERYING Polestar
Z790, CPython 3.11). All tests pass with zero errors.

\textbf{Throughput vs.\ Frame Resolution}: Fast Lane sustains 354K FPS at CartPole
resolution (84$\times$84) and 21K FPS at HD (640$\times$480), far exceeding the 60 FPS
target at every resolution. The blue line shows actual throughput; the dashed
coral line shows what would happen with linear (serialisation-based) scaling.
Sub-linear degradation proves the protocol overhead is $\mathcal{O}(1)$.

\begin{figure}[ht]
\centering
\begin{subfigure}[t]{0.48\textwidth}
  \centering
  \includegraphics[width=\textwidth]{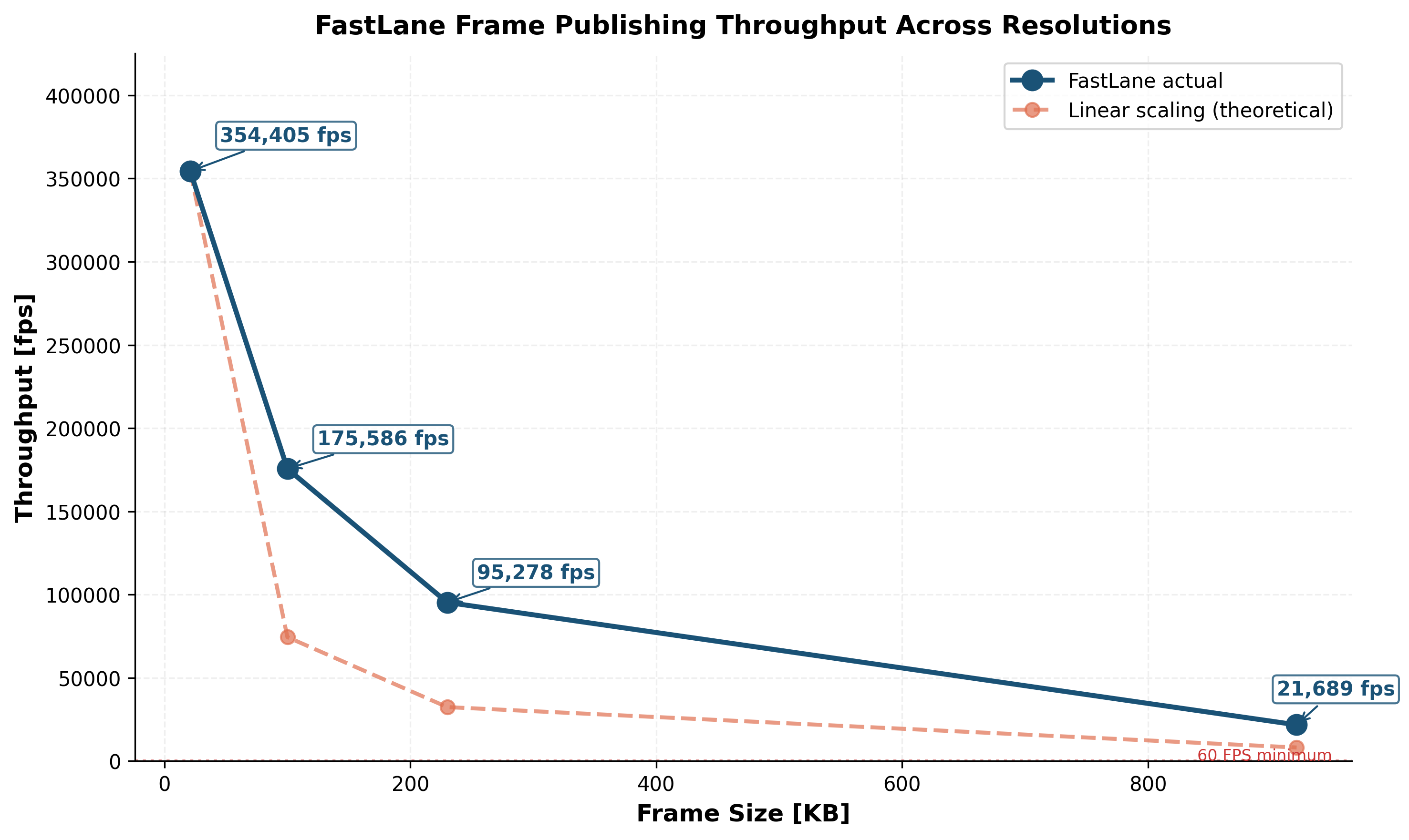}
  \caption{Throughput vs.\ frame resolution.}
  \label{fig:fastlane-throughput}
\end{subfigure}
\hfill
\begin{subfigure}[t]{0.48\textwidth}
  \centering
  \includegraphics[width=\textwidth]{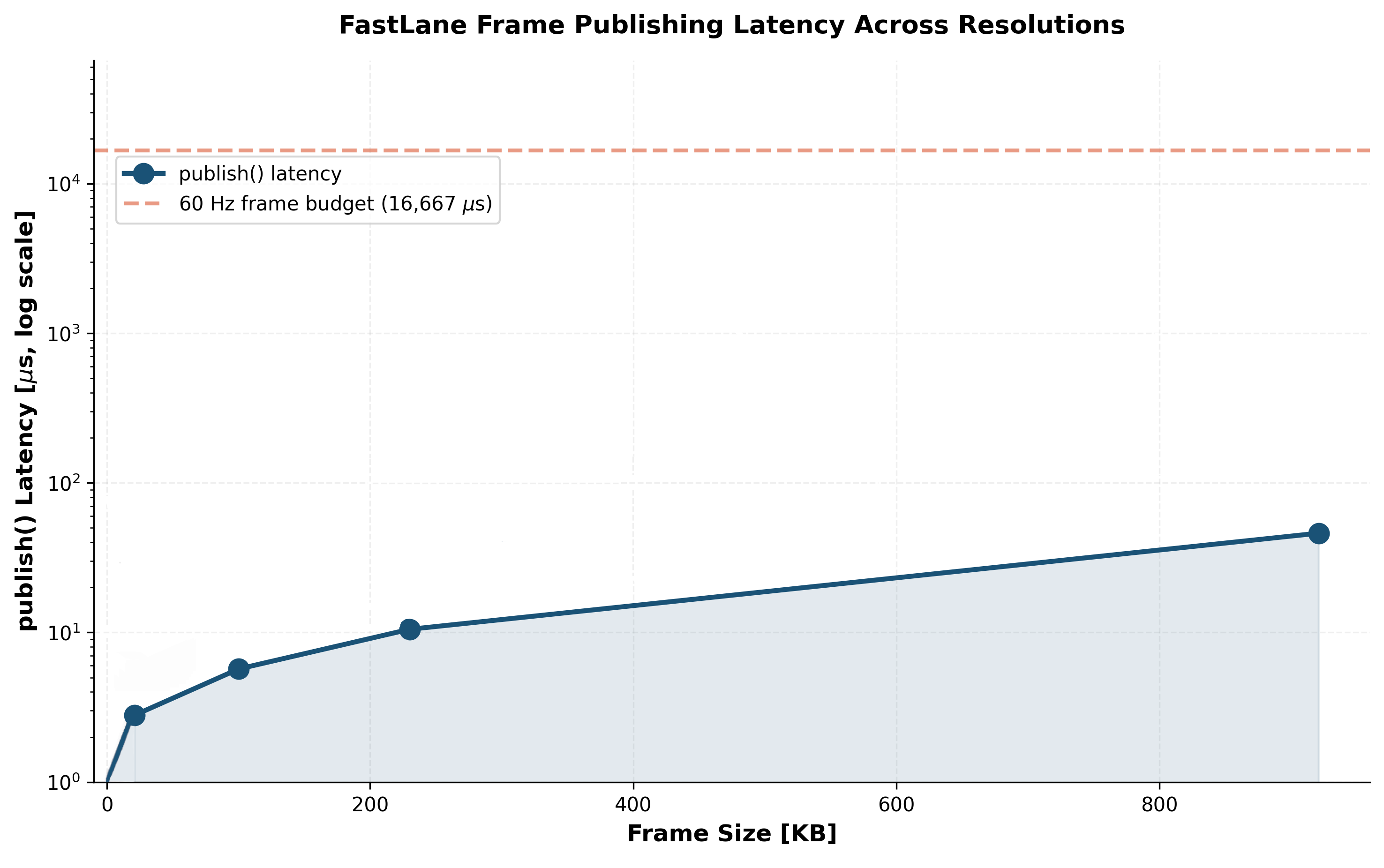}
  \caption{Publish latency vs.\ frame size.}
  \label{fig:fastlane-latency}
\end{subfigure}
\caption{FastLane empirical benchmarks. (a) Throughput reaches 354K FPS at 84$\times$84
and degrades sub-linearly, proving $\mathcal{O}(1)$ protocol overhead. (b) Latency
remains far below the 16,667~$\mu$s budget at every resolution; even 640$\times$480 HD
costs only 46~$\mu$s, giving 362$\times$ headroom at 60 Hz.}
\label{fig:fastlane-bench}
\end{figure}

\textbf{Writer Decoupling from Reader Speed}: Writer throughput is 337K fps with
no reader, 329K fps with a 1 Hz reader, and 328K fps with a 60 Hz reader
(2.5\% variance, within OS scheduling noise). This proves the writer never
waits for the reader. That is the mathematical definition of lock-free streaming.
The 337K ``no reader'' baseline is slightly below the 354K figure above because
the two numbers come from different benchmark files:
\texttt{test\_fastlane\_resolution.py} runs 20,000 frames in a tight loop with no
reader-subprocess machinery around it, while \texttt{test\_fastlane\_decoupling.py}
runs 50,000 frames under the multi-process harness with \texttt{capacity=64} in
every condition (including ``no reader''), which changes cache footprint and
scheduling. Both are correct measurements of what they claim; they are not
two attempts at the same measurement.

\begin{figure}[ht]
\centering
\includegraphics[width=\textwidth]{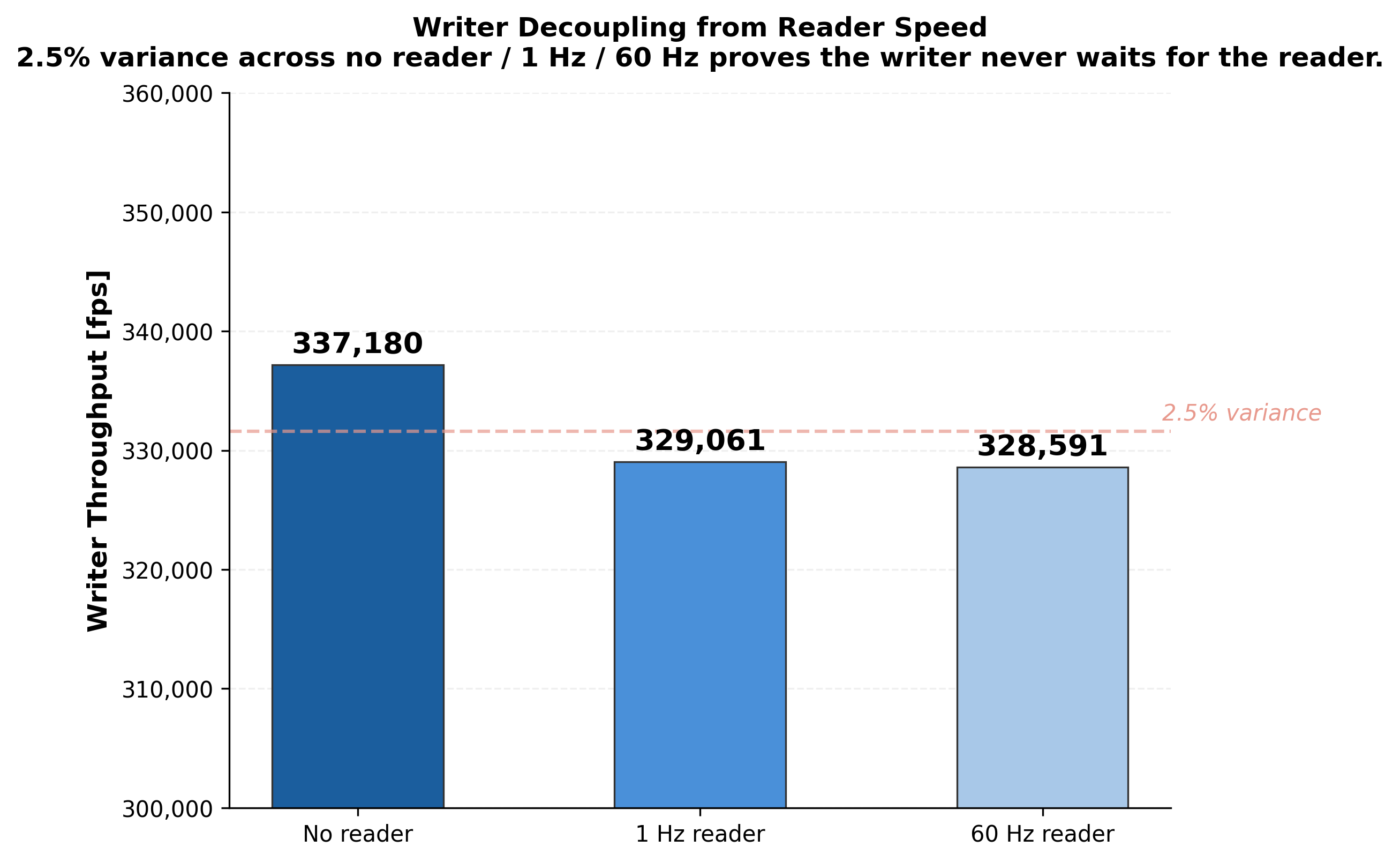}
\caption{FastLane writer decoupling proof.}
\label{fig:fastlane-decoupling}
\end{figure}

\textbf{Publish Latency vs.\ Frame Size}: All resolutions remain far below the
16,667~$\mu$s budget required for 60 Hz (dashed line). Even at 640$\times$480 HD,
publish latency is 46~$\mu$s, which is 362$\times$ faster than needed. The log scale
reveals the massive headroom at every resolution.

\begin{table}[ht]
\centering
\begin{tabular}{@{}lll@{}}
\toprule
\textbf{Metric} & \textbf{Value} & \textbf{Condition} \\
\midrule
Torn reads                        & 0 / 155,000 frames        & Concurrent writer + reader \\
Writer throughput variance        & 2.5\%                     & No reader / 1 Hz / 60 Hz \\
\texttt{publish()} latency p50    & 2.9~$\mu$s                & 84$\times$84 RGB (21 KB) \\
\texttt{publish()} latency p99    & 4.8~$\mu$s                & 84$\times$84 RGB (21 KB) \\
Throughput at HD (640$\times$480) & 21,689 fps                & 921 KB per frame \\
Latency growth vs.\ frame size    & 13.7$\times$ for 44$\times$ size & Sub-linear ($\mathcal{O}(1)$ overhead) \\
Memory ordering errors            & 0 / 700,000 frames        & CPU-affinity pinned, cross-core \\
\bottomrule
\end{tabular}
\caption{Fast Lane validation summary.}
\label{tab:fastlane-validation}
\end{table}

\subsection{Per-Framework FastLane Overhead}

\begin{figure}[ht]
    \centering
    \includegraphics[width=\textwidth]{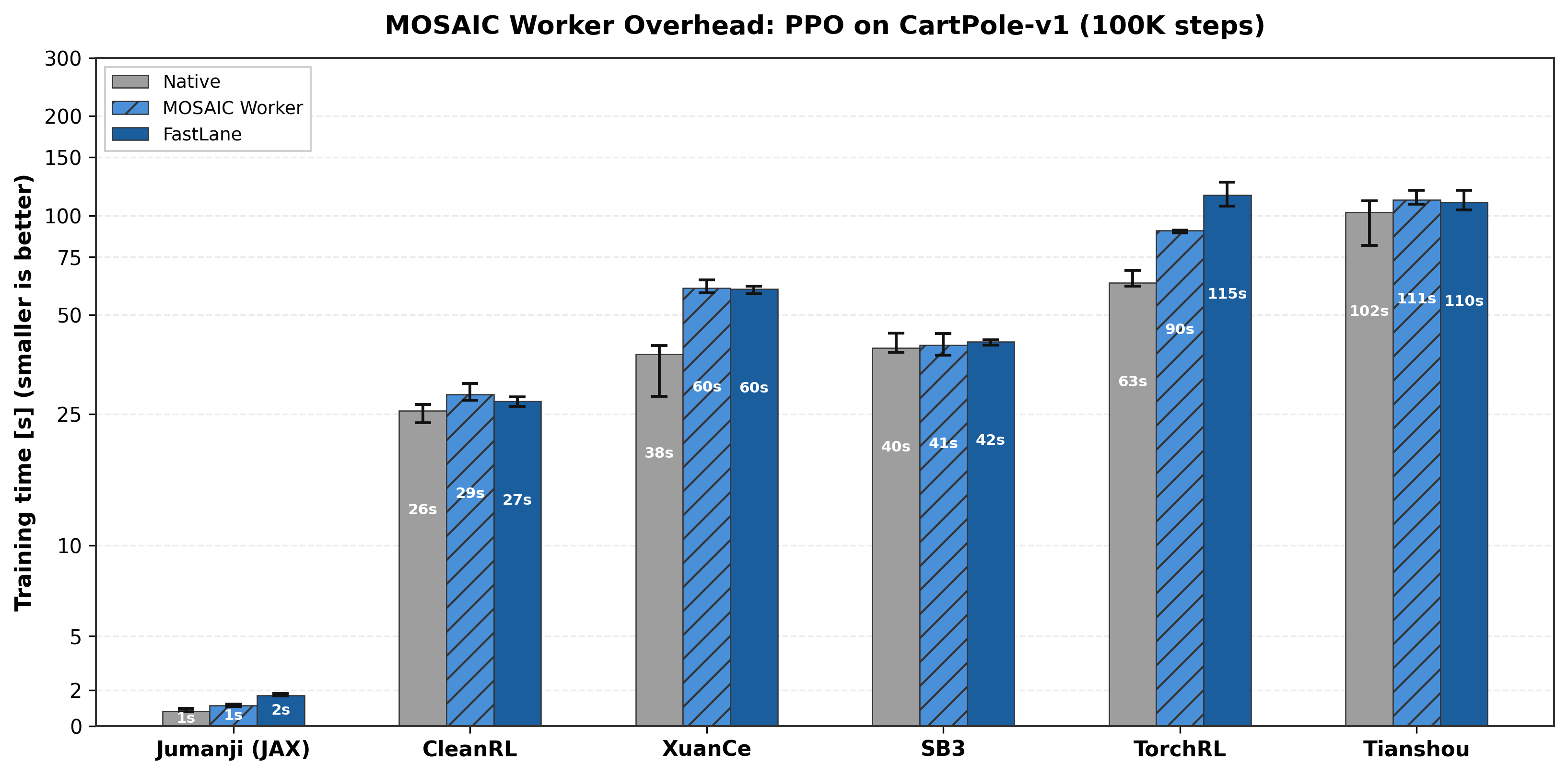}
    \caption{FastLane combined overhead across frameworks.}
    \label{fig:fastlane-combined-overhead}
\end{figure}

Enabling FastLane adds a per-step frame extraction cost inside each worker's
actor loop. The overhead is framework-dependent: frameworks with faster
baseline steps (\eg, SBX's JIT-compiled step) show the largest relative
overhead precisely because their baseline is so fast. The following two plots
summarise the overhead measured across seven publishing frameworks (CleanRL,
SBX, XuanCe, SB3, Tianshou, TorchRL, RLlib) on CartPole-v1, 100,000 steps,
five seeds per condition.

\textbf{Overhead summary}: absolute steps/s comparison between native
(no FastLane), worker-wrapped (subprocess overhead only), and FastLane-enabled
conditions, across all seven frameworks.

\subsection{Limitations}

\textbf{Single-machine constraint.}
FastLane requires both the training worker and the GUI process to run on the
same operating system kernel. POSIX shared memory (\texttt{mmap}) works by mapping
the same physical RAM pages into two process address spaces; across two
machines there is no shared physical RAM and no bridge. This is the same
tradeoff made by Sample Factory~\cite{petrenko2020sample}, which restricts its
shared-memory IPC to single-machine settings in exchange for eliminating
serialisation overhead entirely. Distributed deployments, where workers run on
remote compute nodes, fall back to the Slow Lane's gRPC transport, which works
across any network boundary.

\bibliography{references}

\end{document}